\pdfoutput=1

\documentclass[letterpaper, 10 pt, conference]{ieeeconf}  %

\IEEEoverridecommandlockouts                              %

\usepackage{mathptmx} %
\usepackage{times} %
\usepackage{amsmath} %
\usepackage{amssymb}  %
\usepackage{kotex}
\usepackage[final]{graphicx}
\usepackage{booktabs}
\usepackage{multirow}
\usepackage{xcolor}
\usepackage{colortbl}
\usepackage{capt-of}
\makeatletter\let\IEEElabelindentsave\labelindent\let\labelindent\relax\makeatother
\usepackage{enumitem}
\let\labelindent\IEEElabelindentsave
\usepackage{arydshln}

\usepackage{cite} %
\usepackage{afterpage}
\definecolor{fdrgreen}{HTML}{1C7536}
\makeatletter\let\NAT@parse\undefined\makeatother %
\usepackage[colorlinks=true,linkcolor=fdrgreen,citecolor=fdrgreen,urlcolor=fdrgreen]{hyperref}

\definecolor{header}{RGB}{235,235,235}
\definecolor{ours}{RGB}{226,240,217} %

\newcommand{\edited}[1]{#1}
\newcommand{\editcolor}{}
\newcommand{\editednew}[1]{\textcolor{orange}{#1}}
\renewcommand{\editednew}[1]{#1} %
\newcommand{\editedgrammar}[1]{\textcolor{orange}{#1}}
\renewcommand{\editedgrammar}[1]{#1} %

\title{\LARGE \bf
FlashDexRetarget: Accelerating Dexterous Manipulation \\ Data Generation through Multi-Motion Retargeting

}

\newcommand{\showauthors}{}

\ifdefined\showauthors
\author{Kyungmin Lee$^{1*}$, Sibeen Kim$^{1*}$, Dongyoon Hwang$^{1*}$, Yoonsang Oh$^{1*}$, Donghu Kim$^{2}$, Youngdo Lee$^{2}$,\\
I Made Aswin Nahrendra$^{2}$, Jaegul Choo$^{1\dagger}$, Hojoon Lee$^{2\dagger}$\\[0.3em] %
{\normalsize $^{1}$KAIST AI \qquad $^{2}$Holiday Robotics}\\
{\small \href{https://holiday-robot.github.io/FlashDexRetarget/}
{\texttt{holiday-robot.github.io/FlashDexRetarget}}}}

\else
\author{Anonymous Authors}
\fi

\begin{document}

\twocolumn[{%
  \renewcommand\twocolumn[1][]{#1}%
  \maketitle
  \centering
  \vspace{-3mm} %
  \includegraphics[width=\textwidth]{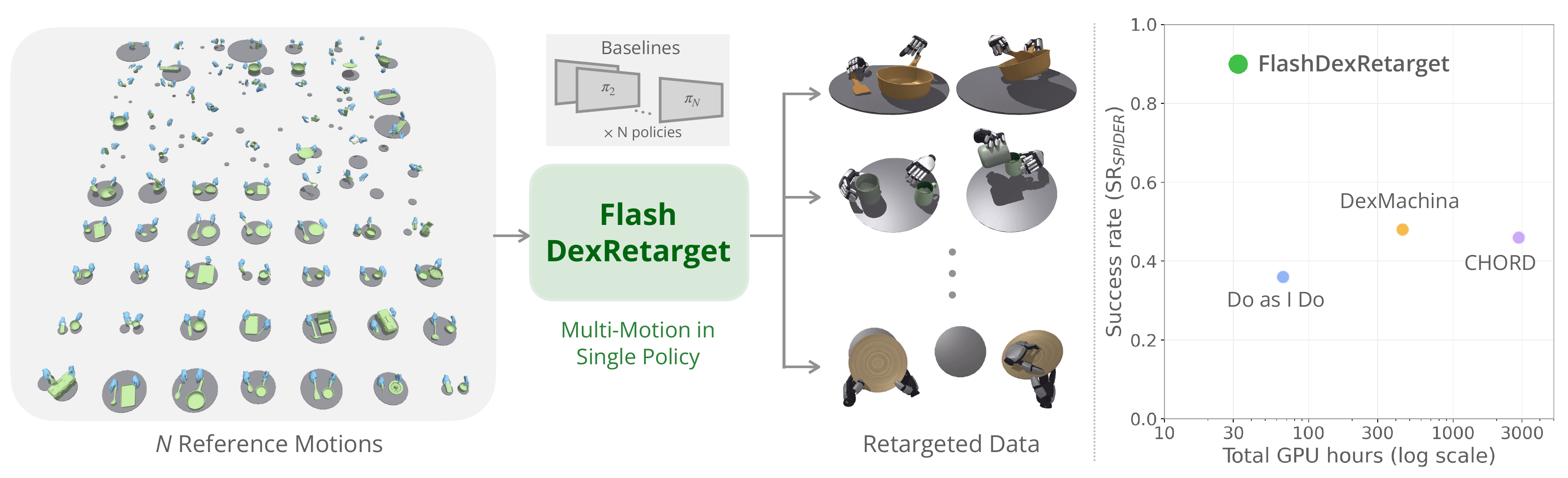}
  \vspace{-6mm} %
  \captionof{figure}{\textbf{FlashDexRetarget at a glance.}
FlashDexRetarget converts collections of human hand-object demonstrations
into physically grounded robot trajectories through multi-reference
RL-based retargeting.
\editedgrammar{Left- and right-hand} policies are jointly trained across reference motions,
avoiding repeated optimization for individual demonstrations, and
successful rollouts are stored as robot data.
The right plot shows cumulative conversion success rate versus total \editedgrammar{GPU-hours}
on a logarithmic scale.
CHORD results are obtained with its released implementation, which \editedgrammar{trains policies} with PPO~\cite{schulman2017ppo}.} %
  \label{fig:main_figure}
  \vspace{6mm}
}]

\ifdefined\showauthors
\begingroup
\renewcommand{\thefootnote}{}
\footnotetext{$^{*}$Equal contribution. \edited{$^{\dagger}$Corresponding author.}}
\endgroup
\fi

\thispagestyle{empty}
\pagestyle{empty}

\afterpage{\afterpage{%
\begin{figure*}[t]
  \centering
  \setlength{\abovecaptionskip}{1.1pt} %
  \includegraphics[width=\textwidth]{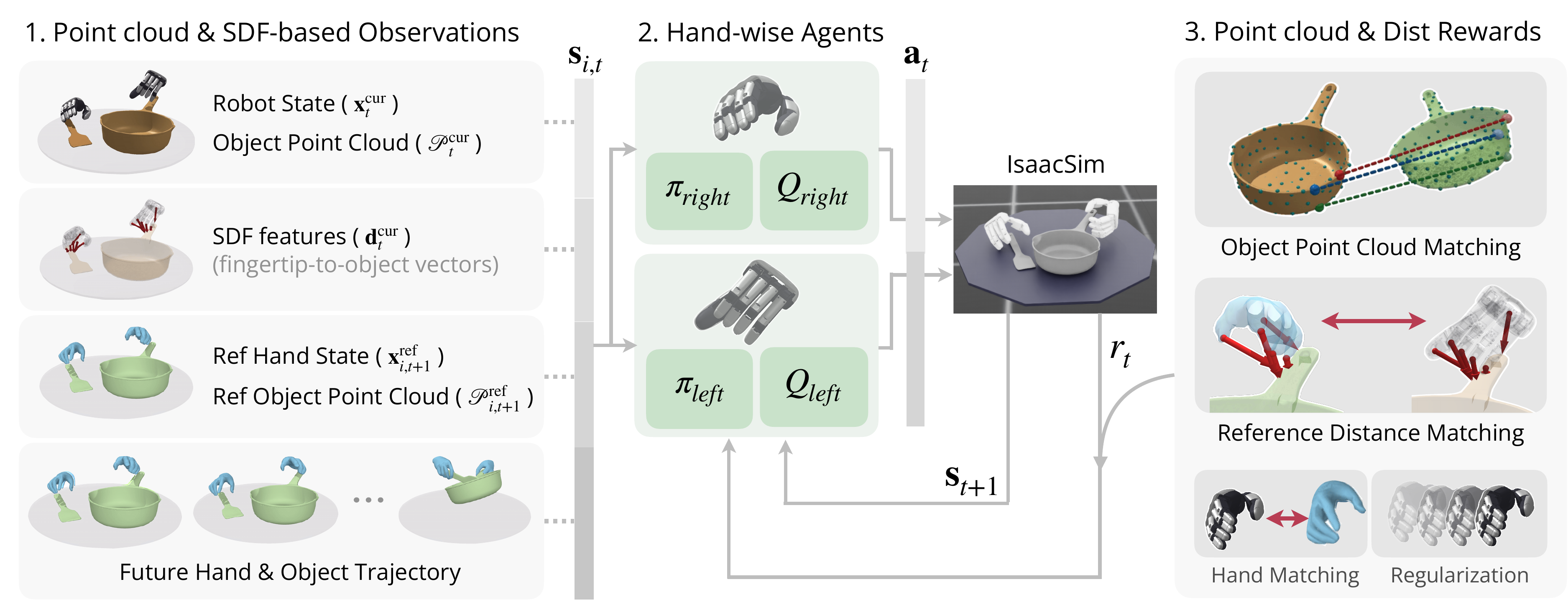}
  \caption{\textbf{Overview of FlashDexRetarget.}
  (1) Observations combine robot states, current and reference object point
  clouds, local signed-distance features, and an encoding of future reference
  motion.
  (2) \editedgrammar{Left- and right-hand} actor-critic pairs are trained with FlashSAC and jointly
  control the simulated interaction, with each pair sharing its parameters across
  all reference motions.
  (3) Object surface alignment and reference-conditioned fingertip proximity
  provide geometric learning signals, complemented by hand tracking and
  regularization. Together, these components support learning across diverse
  interactions.} %
  \label{fig:overview}
\end{figure*}
}}

\begin{abstract}
Human hand-object demonstrations provide a scalable source of data for dexterous robot learning, but transferring them across embodiments requires physically feasible retargeting. Existing physics-based methods typically optimize each demonstration independently, leading to \editedgrammar{either} limited success under finite simulation budgets or training costs that grow with dataset size. We introduce \textbf{FlashDexRetarget}, an RL framework for multi-reference dexterous retargeting. We formulate retargeting as multi-reference tracking, jointly learning a single policy across many demonstrations with off-policy RL and geometric supervision of the demonstrated interactions. This shared training formulation amortizes optimization across references while enabling the policy to track diverse hand-object interactions. On a 50-motion benchmark from TACO, OakInk2, and HOT3D using XHand and Sharpa Wave Hand as target embodiments, FlashDexRetarget retargets 90\% of demonstrations using about 30 GPU-hours, compared with about 46\% at about 3,000 GPU-hours for CHORD. This corresponds to about 100 times lower training compute and a 44-percentage-point improvement in retargeting success. Ablations examine the key design choices, while experiments with up to 1,000 motions and real-world replay further demonstrate the scalability and practical applicability of our method.
\end{abstract}

\section{Introduction}

Learning dexterous manipulation requires large amounts of robot interaction data.
Human hand-object demonstrations~\cite{liu2024taco,zhan2024oakink2,banerjee2025hot3d} provide a promising source of such data, as diverse manipulation behaviors can be collected at scale without directly operating a robot.
To use these demonstrations for robot learning, however, human motions must be translated into actions executable by a target robot hand.
Physics-based retargeting addresses this issue by adapting human demonstrations to the morphology and actuation of the robot in simulation, while leveraging physical contact to reproduce the demonstrated object motion.
To make large human motion datasets practical sources of robot data, retargeting must achieve high success at low compute cost.

Existing physics-based retargeting methods adopt a single-reference formulation, optimizing each demonstration independently.
Sampling-based methods~\cite{paliwal2026dexterous,pan2025spider} search directly over action sequences through simulated rollouts, but high-dimensional actions and long contact sequences make this search challenging under a finite simulation budget.
RL-based methods~\cite{zhu2026chord,li2025maniptrans,mandi2025dexmachina} instead learn closed-loop policies to track the demonstrated hand-object interaction, but train a separate policy for each reference.
Despite their \editedgrammar{different optimization strategies}, both approaches retarget each demonstration independently, causing computational cost to grow linearly with the number of demonstrations.

To avoid this repeated optimization, we study \emph{multi-reference tracking}, where a single reference-conditioned policy is trained jointly across many demonstrations.
By sharing learning across references, this formulation has the potential to amortize training cost over an entire dataset while retaining accurate tracking.
Multi-reference tracking has been effective for humanoid whole-body motion~\cite{luo2026sonic}, but dexterous manipulation introduces additional challenges.

First, object shapes and hand-object configurations vary across demonstrations, requiring the policy to distinguish interaction geometries. Second, reconstructed human demonstrations, even when captured using motion-capture equipment, contain contact errors, making exact contact matching unreliable. Third, multi-reference training broadens the state distribution, increasing the need for model capacity and effective experience reuse. Finally, for bimanual motions, the two hands assume distinct roles, such as primary and supporting roles, requiring asymmetric learning signals.

To address these challenges, we introduce \textbf{FlashDexRetarget}, an RL
framework for multi-reference dexterous retargeting.
First, we condition the policy on object geometry and $K$ future reference frames to distinguish diverse interactions across references.
Second, we replace exact contact matching with distance-based interaction
rewards to provide dense hand-object supervision despite imperfect and sparse
reconstructed contact information.
Third, we adopt the off-policy learner FlashSAC~\cite{kim2026flashsac} and
increase both the replay buffer size and critic capacity to support learning across a broader
multi-reference distribution.
Finally, we use separate left- and right-hand actor-critic networks to
handle the distinct roles and learning signals of the two hands.

We evaluate FlashDexRetarget on 50 hand-object motions from TACO, OakInk2, and HOT3D~\cite{liu2024taco,zhan2024oakink2,banerjee2025hot3d} using XHand and Sharpa Wave Hand as target embodiments.
Against sampling-based and RL-based baselines, we evaluate retargeting success, measured as the fraction of successfully retargeted demonstrations, together with the total training compute required.
A single FlashDexRetarget policy retargets 90\% of the benchmark using about 30 GPU-hours, compared with about 46\% success at about 3,000 GPU-hours for CHORD~\cite{zhu2026chord}.
This corresponds to approximately $100\times$ lower training compute and a \edited{44-percentage-point} improvement in retargeting success.
To understand these gains, we ablate how conditioning on $K$ future reference frames, hand-object interaction objectives, off-policy training and capacity scaling, and separate left- and right-hand networks contribute to performance in the multi-reference setting.
We further show that the number of successfully retargeted trajectories continues to grow as the dataset scales.
Finally, we validate the retargeted trajectories through real-world replay.

Our main contributions are: 
\begin{enumerate}
    \item[(i)] We introduce FlashDexRetarget, an RL framework that jointly retargets human hand-object demonstrations through multi-reference tracking with a single policy.
    \item[(ii)] We combine object point-cloud observations, hand-object interaction objectives, off-policy learning with larger replay and model capacities, and \editedgrammar{left- and right-hand} actor-critic networks for joint dexterous tracking.
    \item[(iii)] We evaluate on 50 motions and two robot embodiments, achieving approximately $100\times$ lower training compute and \edited{44} percentage points higher success than CHORD, with ablations and real-world replay.
\end{enumerate}

\section{Related Work}
\subsection{Human Hand-Object Interaction Data}
Human hand-object datasets cover diverse objects and manipulation behaviors.
DexYCB provides annotated hand-object poses~\cite{chao2021dexycb}, while GRAB and OakInk capture grasping configurations and interaction geometry~\cite{taheri2020grab,yang2022oakink}.
HOI4D and HOT3D extend this coverage to egocentric interactions~\cite{liu2022hoi4d,banerjee2025hot3d}, and TACO, OakInk2, and ARCTIC capture coordinated bimanual manipulation~\cite{liu2024taco,zhan2024oakink2,fan2023arctic}.
GigaHands offers large-scale recordings of activities such as pouring, assembling objects, and playing cards~\cite{fu2025gigahands}.
These datasets provide diverse manipulation experience for robot learning.

However, these recordings capture human hands, \editedgrammar{which a robot cannot execute directly. Kinematic retargeting is a common first step to close this embodiment gap.} UniDex~\cite{zhang2026unidex} converts human hand-object trajectories into robot trajectories by solving fingertip inverse kinematics and \editedgrammar{manually adjusting the hand base} so that the retargeted fingers still appear to touch the object.
VideoManip~\cite{chen2026dexterous} retargets human hand-object trajectories through keypoint matching and repairs the resulting contacts with contact optimization. In both, contact is imposed geometrically after the fact rather than produced by the robot hand acting on the object, so it is not guaranteed that executing the trajectory reproduces the demonstrated object motion. Retargeting must therefore be physics-based, \edited{producing trajectories with verified contacts.}

\subsection{Dexterous Retargeting from Human Demonstrations}
Kinematic and planning-based pipelines adapt human demonstrations to robot motions~\cite{qin2023anyteleop,lakshmipathy2025kinematic,mu2026deximit,xin2026analyzing,Handa-2020-126758}, while physics-based optimization additionally accounts for hand-object contact dynamics in simulation~\cite{liu2024parameterized,yang2025physicsgen}.
Within this physics-based setting, sampling-based methods search over simulated control rollouts~\cite{deboer2005cem,williams2017mppi,howell2022predictivesampling} to support dexterous planning~\cite{li2025drop} and demonstration refinement~\cite{si2025exostart}.
For demonstration retargeting, SPIDER~\cite{pan2025spider} introduces temporary virtual forces between robot fingers and the object to guide annealed sampling toward trajectories that reproduce the demonstrated contacts.
Building on this approach, Do as I Do~\cite{paliwal2026dexterous} incorporates hand-object reconstruction and physics-based refinement to retarget video demonstrations.
Although these methods avoid training a tracking policy, they require a separate trajectory search for each reference, with effectiveness dependent on initialization and reference quality.

Rather than searching trajectories directly, RL-based methods learn dexterous manipulation policies from human motion references and task objectives~\cite{qin2022dexmv,chen2025objdex,li2025maniptrans,mandi2025dexmachina,lum2025human2sim2robot,yuan2025hermes,zhu2026chord,rajeswaran2017learning,feng2026minimalist}.
ManipTrans~\cite{li2025maniptrans} learns residual corrections to a pretrained hand-motion imitator, while DexMachina~\cite{mandi2025dexmachina} progressively reduces object assistance during training.
CHORD~\cite{zhu2026chord} instead uses contact-wrench guidance to align human and robot object-motion capabilities.
Although these methods replace trajectory search with policy learning, they still require separate training or adaptation for each demonstration.
In contrast, FlashDexRetarget amortizes this cost by jointly learning hand-object tracking across demonstrations.

\section{Method}

Our goal is to efficiently generate a large number of physically
grounded robot trajectories from an extensive collection of human
\editedgrammar{hand-object} demonstrations.

Given a dataset of reference \editedgrammar{hand-object} trajectories,
\begin{equation}
    \mathcal{D}
    =
    \{\tau_i^{\mathrm{ref}}\}_{i=1}^{N},
    \qquad
    \tau_i^{\mathrm{ref}}
    =
    \{\mathbf{s}_{i,t}^{\mathrm{ref}}\}_{t=1}^{H_i},
\end{equation}
where $i$ indexes the reference demonstration,
$\mathbf{s}_{i,t}^{\mathrm{ref}}$ contains the recorded human hand states
and object poses at time $t$, and $H_i$ denotes the trajectory length,
\edited{the goal is to reproduce the demonstrated interaction on a target robot.}

Unlike single-reference retargeting, where a separate policy is optimized
for each demonstration, we jointly train a reference-conditioned policy
over the entire collection $\mathcal{D}$.
During training, reference trajectories are sampled from $\mathcal{D}$ and
assigned to parallel simulation environments.
At each step, the policy receives the current simulated state together
with the next reference state:
\begin{equation}
    \mathbf{s}_{i,t}
    =
    \bigl(
        \mathbf{s}^{\mathrm{cur}}_t,\,
        \mathbf{s}^{\mathrm{ref}}_{i,t+1}
    \bigr).
\end{equation}
A policy
$\pi_{\theta}(\mathbf{a}_t \mid \mathbf{s}_{i,t})$
predicts a robot action $\mathbf{a}_t$ to track the reference object
motion and \editedgrammar{hand-object} interaction.
The policy is optimized using tracking rewards defined from the resulting
robot and object states.
By sharing policy learning across demonstrations, our formulation enables
efficient conversion of large reference collections.

FlashDexRetarget combines \editedgrammar{geometry- and interaction-aware observations with conditioning on $K$ future reference frames} (Section~\ref{sec:observations}), object surface tracking and \editedgrammar{hand-object} distance rewards
(Section~\ref{sec:rewards}), hand-wise \editedgrammar{actor-critic} networks for \editedgrammar{the} left and right hands (Section~\ref{sec:architecture}), and off-policy learning
(Section~\ref{sec:learning}).

\subsection{Observation}
\label{sec:observations}

To track multiple references with a shared policy, we provide a
current-state input $\mathbf{s}^{\mathrm{cur}}_t$ and a reference input
$\mathbf{s}^{\mathrm{ref}}_{i,t+1}$. The latter includes the next target
state and motion information over the next $K$ steps:
\begin{equation}
    \begin{aligned}
    \mathbf{s}^{\mathrm{cur}}_t
    &=
    \bigl(
        \mathbf{x}^{\mathrm{cur}}_t,\,
        \mathcal{P}^{\mathrm{cur}}_t,\,
        \mathbf{d}^{\mathrm{cur}}_t
    \bigr), \\    
    \mathbf{s}^{\mathrm{ref}}_{i,t+1}
    &=
    \bigl(
        \mathbf{x}^{\mathrm{ref}}_{i,t+1},\,
        \mathcal{P}^{\mathrm{ref}}_{i,t+1},\,
        \mathbf{z}^{\mathrm{ref}}_{i,t+1}
    \bigr)\editedgrammar{,}    
    \end{aligned}
    \label{eq:reference_observation}
\end{equation}
where $\mathbf{x}^{\mathrm{cur}}_t$ contains the current robot and object
states, and $\mathbf{x}^{\mathrm{ref}}_{i,t+1}$ contains the reference
hand-object states at time $t+1$.

\paragraph{Interaction-aware observations}
To explicitly represent object geometry and hand-object relationships
across references, we use object point clouds and \editednew{signed distance field (SDF)-based} hand-object distance
features.
We express these features in the wrist-local frame to provide a consistent
hand-centered representation across different hand poses and object motions.
Specifically, we represent the object surface using 128 uniformly sampled
points.
The point clouds at the current simulated and next reference poses are
$\mathcal{P}^{\mathrm{cur}}_t$ and
$\mathcal{P}^{\mathrm{ref}}_{i,t+1}$, respectively.
Furthermore, to describe local hand-object relationships, we compute distances
from each fingertip and wrist to the object surface, together with the
corresponding surface normal vectors in the wrist frame.
We concatenate these distances and normals to form
$\mathbf{d}^{\mathrm{cur}}_t$.
\editednew{Here, the distances are queried from a precomputed SDF of each object, discretized on a $128^3$ voxel grid covering a $60\,\mathrm{cm}$ cube centered at the object origin, since computing signed distances directly from the object mesh at every simulation step is computationally expensive.}

\paragraph{$K$ \texorpdfstring{\editedgrammar{future}}{future} reference conditioning}
Most prior dexterous retargeting methods condition the policy solely on the next reference frame~\cite{li2025maniptrans,mandi2025dexmachina,zhu2026chord}, providing limited temporal context for anticipating subsequent motion.
Prior locomotion tracking methods have shown that conditioning on multiple future reference frames over a short temporal horizon can provide useful motion context~\cite{han2025kungfubot2,lee2025phuma}.
We therefore encode the reference hand and object states from $t+1$ to
$t+K$, represented relative to the wrist frame at time $t$:
\begin{equation}
    \mathbf{z}^{\mathrm{ref}}_{i,t+1}
    =
    \mathrm{Enc}_{\psi}
    \left(
        \mathbf{x}^{\mathrm{ref}}_{i,t+1:t+K}
    \right),
    \label{eq:future_reference}
\end{equation}
where $K=10$ and $\mathrm{Enc}_{\psi}$  is a temporal encoder that maps the
future reference sequence to a 128-dimensional latent representation. The encoder is trained from scratch jointly with the actor and critic.
The policy is thus conditioned on both the next reference state and the future latent, combining immediate tracking targets with broader temporal context.

\subsection{Reward}
\label{sec:rewards}

For each hand $h \in \{\mathrm{left},\mathrm{right}\}$, we define a hand-specific reward as a weighted combination of object tracking, hand-object interaction, hand tracking, and regularization terms:
\begin{equation}
    r_t^h
    =
    \editednew{w_{\mathrm{obj}}}\,r_t^{\mathrm{obj},h}
    +
    \editednew{w_{\mathrm{int}}}\,r_t^{\mathrm{int},h}
    +
    \editednew{w_{\mathrm{track}}}\,r_t^{\mathrm{track},h}
    +
    \editednew{w_{\mathrm{reg}}}\,r_t^{\mathrm{reg},h}.
    \label{eq:total_reward}
\end{equation}

\paragraph{Object tracking reward}
Existing methods~\cite{li2025maniptrans, mandi2025dexmachina} typically
combine object translation and rotation errors using manually chosen
weights. Such weighting requires balancing errors with different units
and scales.
Instead, we measure object tracking using corresponding surface points. We reuse the 128 object-frame surface points from \editedgrammar{Section}~\ref{sec:observations}, $\mathcal{P}^{\mathrm{obj},h}=\{\mathbf{p}_m^h\}_{m=1}^{128}$, and normalize them to a common radius $\rho$ to make the reward comparable across object sizes. We then transform the normalized points using the current and reference object poses to obtain
$\tilde{\mathcal{P}}^{\mathrm{cur},h}$ and $\tilde{\mathcal{P}}^{\mathrm{ref},h}$ in the world frame.
We set $\rho=0.058\,\mathrm{m}$ so that a $30^\circ$ rotation corresponds to approximately $3\,\mathrm{cm}$ of point displacement, consistent with the success thresholds~\cite{li2025maniptrans}.
To emphasize the largest object misalignments, we use the mean of the three largest pointwise errors rather than averaging over all points. Let $\mathcal{I}_{h,t}^{(k)}$ denote the indices of the $k=3$ largest pointwise errors.
We define
\begin{equation} 
    \begin{aligned} 
        \editcolor \tilde{\mathbf{p}}_m^h
        &\editcolor = \rho\,
        \left(
        \frac{\mathbf{p}_m^h}{\max_{n}\left\|\mathbf{p}_n^h\right\|_2}
        \right), \\
        e_t^{\mathrm{obj},h} 
        &= \frac{1}{k} \sum_{m\in\mathcal{I}_{h,t}^{(k)}} \left\| {\editcolor \tilde{\mathcal{P}}^{\mathrm{cur},h}_{t+1,m}}-{\editcolor \tilde{\mathcal{P}}^{\mathrm{ref},h}_{i,t+1,m}} \right\|_2, \\ 
        r_t^{\mathrm{obj},h} 
        &= \exp\!\left( -e_t^{\mathrm{obj},h}/ \sigma_{\mathrm{obj}} \right), 
    \end{aligned} 
    \label{eq:object_tracking_reward} 
\end{equation}
where $\sigma_{\mathrm{obj}}=0.1\,\mathrm{m}$.
Thus, the metric measures translation and rotation in a common unit without separate weights.

\paragraph{Hand-object interaction reward}
\editedgrammar{Prior work}~\cite{mandi2025dexmachina,zhu2026chord} \editedgrammar{infers} contact using a fixed \editedgrammar{hand-object} distance threshold.
However, reconstruction gaps can exceed this threshold, causing demonstrated contacts to be incorrectly labeled as non-contact.
We instead use continuous \editedgrammar{hand-object} distances to supervise proximity without assigning binary contact labels.
For the robot hand, we reuse the SDF-based distances introduced in \editedgrammar{Section}~\ref{sec:observations}, retaining only the five fingertip distances and excluding the wrist distance.
Let $d_{h,t}^{(f)}$ denote the signed distance from robot fingertip $f$ to the object surface.
For the corresponding reference fingertip, we compute the unsigned distance $d_{h,t}^{\mathrm{ref},(f)}$ to the nearest vertex of the object mesh.
Using unsigned reference distances avoids requiring the robot to reproduce penetrations present in the reconstructed demonstration.
Rather than matching the reference distance exactly, we penalize only the amount by which the robot distance exceeds it:
\begin{equation}
    \begin{aligned}
        \epsilon_{h,t}^{(f)}
        &= \max\!\left(0,\ d_{h,t}^{(f)} - d_{h,t}^{\mathrm{ref},(f)}\right), \\
        r_t^{\mathrm{int},h}
        &= \frac{1}{5}\sum_{f=1}^{5}
        \exp\!\left(-\epsilon_{h,t}^{(f)}/\sigma_{\mathrm{int}}\right),
    \end{aligned}
    \label{eq:interaction_reward}
\end{equation}
where $\sigma_{\mathrm{int}}=0.01\,\mathrm{m}$.
Each fingertip receives the maximum reward when
$d_{h,t}^{(f)} \leq d_{h,t}^{\mathrm{ref},(f)}$, with the reward decreasing as the excess increases.
\edited{\editednew{This allows the robot to close} reconstruction gaps rather than reproduce them.}

\paragraph{Hand tracking reward}
\editednew{To keep the robot hand motion close to the demonstration and exploit human manipulation priors, we follow prior work~\cite{li2025maniptrans,mandi2025dexmachina} and include a hand tracking term $r_t^{\mathrm{track},h}$.
It combines exponential rewards on wrist translation, wrist rotation, and fingertip position errors with a behavior-cloning term that encourages commanded joint positions to match kinematic retargets.}

\paragraph{Regularization}
\editednew{To keep the resulting robot motions physically executable, we penalize large actions $\mathbf{a}_t^h$ in $r_t^{\mathrm{reg},h}$.
We also penalize tracking failures, which occur when the object or any fingertip deviates from its reference position by more than a predefined threshold. \editedgrammar{Terminations due to timeouts or successful clip completion are not penalized.}}

\subsection{Bimanual Architecture}
\label{sec:architecture}
\editednew{Prior RL-based retargeting methods~\cite{li2025maniptrans,mandi2025dexmachina,zhu2026chord} control both hands with a single \editedgrammar{actor-critic} pair trained on their combined reward, which makes it difficult to attribute success or failure to either hand.}
\editednew{We therefore assign each hand a dedicated \editedgrammar{actor-critic} pair trained with its own reward $r_t^h$.}
For each hand $h\in\{\mathrm{left},\mathrm{right}\}$, we maintain an actor
$\pi_{\theta_h}$ and critic $Q_{\phi_h}$.
Although the rewards are hand-specific, each actor and critic receives the full
observation $\mathbf{s}_{i,t}$, including the states of both hands. This allows each hand to incorporate cross-hand interactions, such as when the object manipulated by one hand contacts or constrains the object manipulated by the other.
Despite receiving the full observation, each actor outputs only the action of its corresponding hand:
\begin{equation}
    \mathbf{a}_t^{h}
    \sim
    \pi_{\theta_h}\!\left(\cdot\mid \mathbf{s}_{i,t}\right),
    \qquad
    \mathbf{a}_t
    =
    \bigl(
        \mathbf{a}_t^{\mathrm{left}},
        \mathbf{a}_t^{\mathrm{right}}
    \bigr).
    \label{eq:bimanual_actions}
\end{equation}

\begin{figure*}[t]
  \centering
  \includegraphics[width=\textwidth]{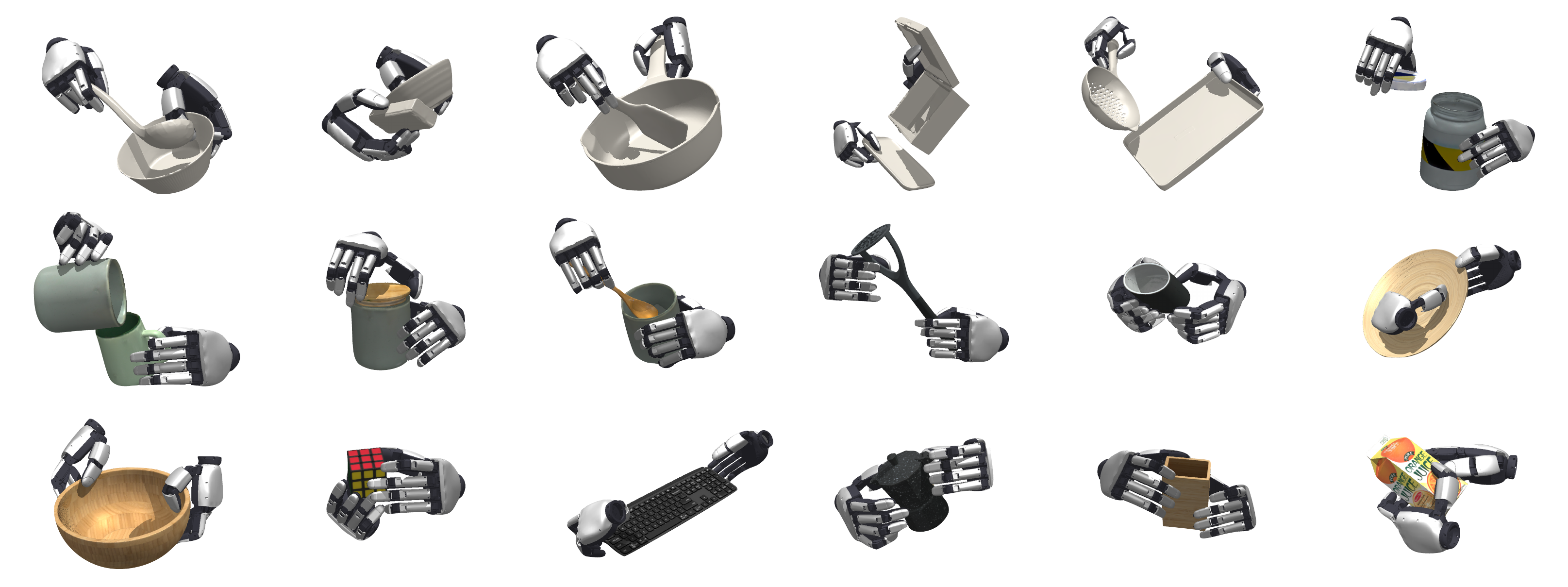}
  \caption{\textbf{Qualitative retargeting results.}
  Representative snapshots of dexterous robot manipulation generated by FlashDexRetarget in simulation\edited{, spanning tool use, container handling, and everyday object manipulation across diverse object geometries and hand-object configurations.}}
  \label{fig:qualitative_results}
\end{figure*}

\subsection{Off-\texorpdfstring{\editedgrammar{Policy}}{Policy} Learning}
\label{sec:learning}
\editednew{On-policy algorithms such as PPO~\cite{schulman2017ppo} discard transitions after each policy update, which is data-inefficient, whereas off-policy algorithms reuse past transitions stored in a replay buffer.
We therefore adopt FlashSAC~\cite{kim2026flashsac}, an off-policy RL algorithm with high sample efficiency.
Unlike single-reference training, however, multi-reference training involves diverse objects as well as hand motions, which yields a much broader state distribution.
To accommodate this, we enlarge the replay buffer from the default $10$M to $50$M transitions to retain experience across references, \editedgrammar{and increase} the critic hidden dimension from the default $256$ to $1024$ to better \editedgrammar{estimate values over} such diverse data.}

\section{Experiments}
\label{sec:experiments}

We evaluate FlashDexRetarget in terms of retargeting success,
computational efficiency, scalability, and real-world executability
through four research questions:\\
\textbf{RQ1:} How does FlashDexRetarget compare with prior methods in
retargeting success and computational efficiency? \\
\textbf{RQ2:} Which design components improve multi-reference retargeting
performance and learning efficiency? \\
\textbf{RQ3:} Does multi-reference retargeting remain effective as the
number of reference motions increases? \\
\textbf{RQ4:} \edited{Do retargeted trajectories transfer to}
real-world robot execution?

\subsection{Experimental Setup}
\label{sec:experimental_setup}

\paragraph{Datasets and preprocessing}
We use human hand-object demonstrations from
TACO~\cite{liu2024taco}, OakInk2~\cite{zhan2024oakink2}, and
HOT3D~\cite{banerjee2025hot3d}, following the preprocessing
procedure of CHORD~\cite{zhu2026chord}.
Our collection includes single-object interactions from HOT3D
and two-object interactions from TACO and OakInk2.
To avoid trivial successes, we exclude inactive clips in which
neither hand performs manipulation and the objects remain
stationary, since such clips can satisfy the tracking criteria
without meaningful robot action.

\paragraph{Evaluation protocol}
For the main comparison, we select 50 reference motions comprising
25 single-object and 25 two-object interactions.
We evaluate retargeting to two robot embodiments, XHand and
Sharpa Wave Hand, and compare FlashDexRetarget with
Do as I Do~\cite{paliwal2026dexterous},
DexMachina~\cite{mandi2025dexmachina}, and
CHORD~\cite{zhu2026chord}.
\edited{We use the released implementations of Do as I Do and CHORD,
where CHORD is trained with PPO~\cite{schulman2017ppo}, and modify
DexMachina to initialize from random states along the reference
trajectory rather than always from the motion start.}
For the scaling experiments, we use reference collections of
200, 500, and 1,000 motions, maintaining approximately balanced proportions of single- and two-object interactions.
\edited{All simulation experiments are conducted in \editedgrammar{Isaac Sim} on NVIDIA RTX 3090 GPUs.}

\begin{table*}[t]
\centering
\caption{Quantitative comparison of dexterous motion retargeting methods}
\vspace{-2mm}
\label{tab:rq1_main}
\setlength{\tabcolsep}{10pt}
\renewcommand{\arraystretch}{1.2}
\begin{tabular}{lccccccc}
\toprule
Method
& $\mathrm{Motion}$
& $\mathrm{SR}_{\mathrm{SPIDER}}$ $\uparrow$
& $\mathrm{SR}_{\mathrm{MT}\text{-}\mathrm{Obj}}$ $\uparrow$
& $\mathrm{SR}_{\mathrm{MT}}$ $\uparrow$
& MOPE (mm) $\downarrow$
& MORE (rad) $\downarrow$
& GPU-h $\downarrow$ \\
\midrule
\multicolumn{8}{l}{\textit{(a) XHand}} \\

\addlinespace[2pt]
Do as I Do~\cite{paliwal2026dexterous}
& Single & 0.36 & 0.04 & 0.00 & 47.32 & 0.28 & 66 \\
DexMachina~\cite{mandi2025dexmachina}
& Single & 0.48 & 0.32 & 0.12 & 37.49 & 0.29 & 447 \\
CHORD~\cite{zhu2026chord}
& Single & 0.46 & 0.02 & 0.00 & 55.70 & \textbf{0.16} & 2,847 \\
\rowcolor{ours}
\textbf{FlashDexRetarget}
& \textbf{Multi} & \textbf{0.90} & \textbf{0.86} & \textbf{0.86}
& \textbf{10.95} & 0.25 & \textbf{29} \\
\midrule
\multicolumn{8}{l}{\textit{(b) Sharpa Wave Hand}} \\
\addlinespace[2pt]
Do as I Do
& Single & 0.64 & 0.10 & 0.10 & 58.33 & 0.24 & 62 \\
CHORD
& Single & 0.50 & 0.22 & 0.00 & 60.07 & \textbf{0.19} & 3,314 \\
\rowcolor{ours}
\textbf{FlashDexRetarget}
& \textbf{Multi} & \textbf{0.86} & \textbf{0.80} & \textbf{0.80}
& \textbf{16.45} & 0.32 & \textbf{44} \\
\bottomrule
\end{tabular}
\end{table*}

\begin{figure*}[t]
  \centering
  \setlength{\abovecaptionskip}{0pt}
  \includegraphics[width=\textwidth]{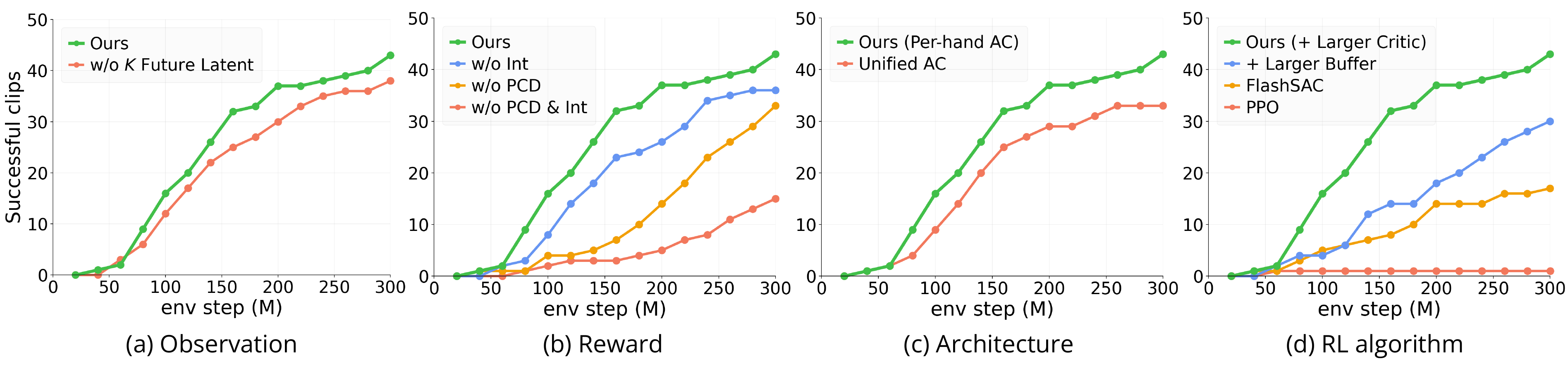}
  \caption{\textbf{Ablation of the components of FlashDexRetarget.}
Curves show the cumulative number of successfully converted reference motions from the 50-motion benchmark over 300 million environment steps. Each reference is counted once a rollout satisfies $\mathrm{SR}_{\mathrm{MT\text{-}Obj}}$.
We examine (a) future reference encoding; (b) point-cloud (PCD) and interaction (Int) reward components, with the corresponding observations and rewards ablated together; (c) a unified \editedgrammar{actor-critic} versus \editedgrammar{left- and right-hand actor-critic} pairs; and (d) PPO, default FlashSAC, and FlashSAC with a larger replay buffer, a wider critic, or both.}
  \label{fig:rq2_ablation}
\end{figure*}

\paragraph{Metrics}
We report three success rates:
$\mathrm{SR}_{\mathrm{MT\text{-}Obj}}$,
$\mathrm{SR}_{\mathrm{MT}}$, and
$\mathrm{SR}_{\mathrm{SPIDER}}$.
The ManipTrans metric
$\mathrm{SR}_{\mathrm{MT}}$~\cite{li2025maniptrans}
uses object position and rotation error thresholds of
$3\,\mathrm{cm}$ and $30^\circ$, together with fingertip and
hand-joint position error thresholds of
$6\,\mathrm{cm}$ and $8\,\mathrm{cm}$, respectively.
Its object-only variant,
$\mathrm{SR}_{\mathrm{MT\text{-}Obj}}$, retains the object-tracking
criteria while omitting the hand-tracking requirements.
\edited{For comparison with prior baselines, we also report
$\mathrm{SR}_{\mathrm{SPIDER}}$~\cite{pan2025spider}, which applies
object position and rotation error thresholds of
$10\,\mathrm{cm}$ and $0.5\,\mathrm{rad}$ to object errors averaged over the
active hands rather than requiring each hand's object to satisfy them.
This averaging can overestimate success when one object
remains nearly stationary in the reference. \editedgrammar{We therefore use
$\mathrm{SR}_{\mathrm{MT\text{-}Obj}}$ for component-wise ablations, as it
measures object retargeting without cross-object averaging, while retaining
$\mathrm{SR}_{\mathrm{SPIDER}}$ for the main comparison with prior methods.}}
To evaluate the quality of successfully retargeted trajectories, we additionally
report the mean object position error (MOPE) and mean object rotation error
(MORE) among trajectories that satisfy the $\mathrm{SR}_{\mathrm{SPIDER}}$
criterion.

\paragraph{Computational cost}
We report computational cost in total GPU-hours.
\edited{For single-reference baselines, we periodically evaluate each
per-motion run, stop it once a successful rollout is obtained and stored, and record
the total GPU-hours consumed up to that point. We sum these costs over all
attempted motions, including the full computational cost of unsuccessful runs.}
For multi-reference training, we use a fixed training-step budget for each experiment and report the total GPU-hours required to reach that budget.
Multi-reference success is measured cumulatively over training: a reference is counted as successful once a rollout satisfying the $\mathrm{SR}_{\mathrm{MT\text{-}Obj}}$ criterion is stored.
The training budgets used for each experiment are specified in the corresponding sections.

\subsection{Retargeting Performance and Computational Efficiency}
\label{sec:performance_comparison}

To evaluate retargeting performance and computational efficiency, we compare
FlashDexRetarget with prior baselines on the 50-motion benchmark.
We train FlashDexRetarget for 300 million environment steps on XHand.
For Sharpa Wave Hand, which has a higher-dimensional control space, we use a slightly larger budget of 400 million steps.
As shown in Table~\ref{tab:rq1_main}, FlashDexRetarget achieves higher
retargeting success while requiring substantially less computation.
\edited{On XHand, under} the $\mathrm{SR}_{\mathrm{SPIDER}}$ criterion, FlashDexRetarget
achieves at least a \edited{40-percentage-point} higher success rate than the
baselines, while using approximately $2\times$ less compute than
Do as I Do and nearly $100\times$ less compute than CHORD.
The difference is more pronounced under the stricter
$\mathrm{SR}_{\mathrm{MT}}$ criterion, which additionally requires the
robot hand to track the reference hand: FlashDexRetarget achieves
\edited{$86\%$} success, compared with $12\%$ for DexMachina and $0\%$ for
Do as I Do and CHORD.
To examine whether these gains depend on the target embodiment, we
additionally evaluate on Sharpa Wave Hand, which is also used by Do as I Do and
CHORD.
The same trend holds. FlashDexRetarget achieves $86\%$ under $\mathrm{SR}_{\mathrm{SPIDER}}$,
compared with $64\%$ for Do as I Do and $50\%$ for CHORD, while requiring
less computation.
Under the stricter $\mathrm{SR}_{\mathrm{MT}}$ criterion,
FlashDexRetarget achieves $80\%$ success, compared with $10\%$ for
Do as I Do and $0\%$ for CHORD.
These results show that the improvement extends across both robot embodiments.

\subsection{Key Components for Multi-Reference Retargeting}
\label{sec:ablation}

We ablate the observation design, rewards,
\editedgrammar{actor-critic} architecture, and training configuration of FlashDexRetarget.
All variants are trained for 300 million environment steps.
Figure~\ref{fig:rq2_ablation} shows their learning progress and
cumulative retargeting success measured by
$\mathrm{SR}_{\mathrm{MT\text{-}Obj}}$.

\paragraph{$K$ \texorpdfstring{\editedgrammar{future}}{future} reference conditioning}
To evaluate the importance of $K$ future reference conditioning, we remove the future trajectory encoding while retaining the next-step reference input.
\edited{Compared with this variant,} FlashDexRetarget
with future reference conditioning accumulates successful trajectories
faster and converts more references by the end of training.
This demonstrates that temporal context beyond the immediate target helps the policy anticipate subsequent motion, improving hand pose and object rotation tracking.

\begin{figure*}[t]
  \centering
  \includegraphics[width=\textwidth]{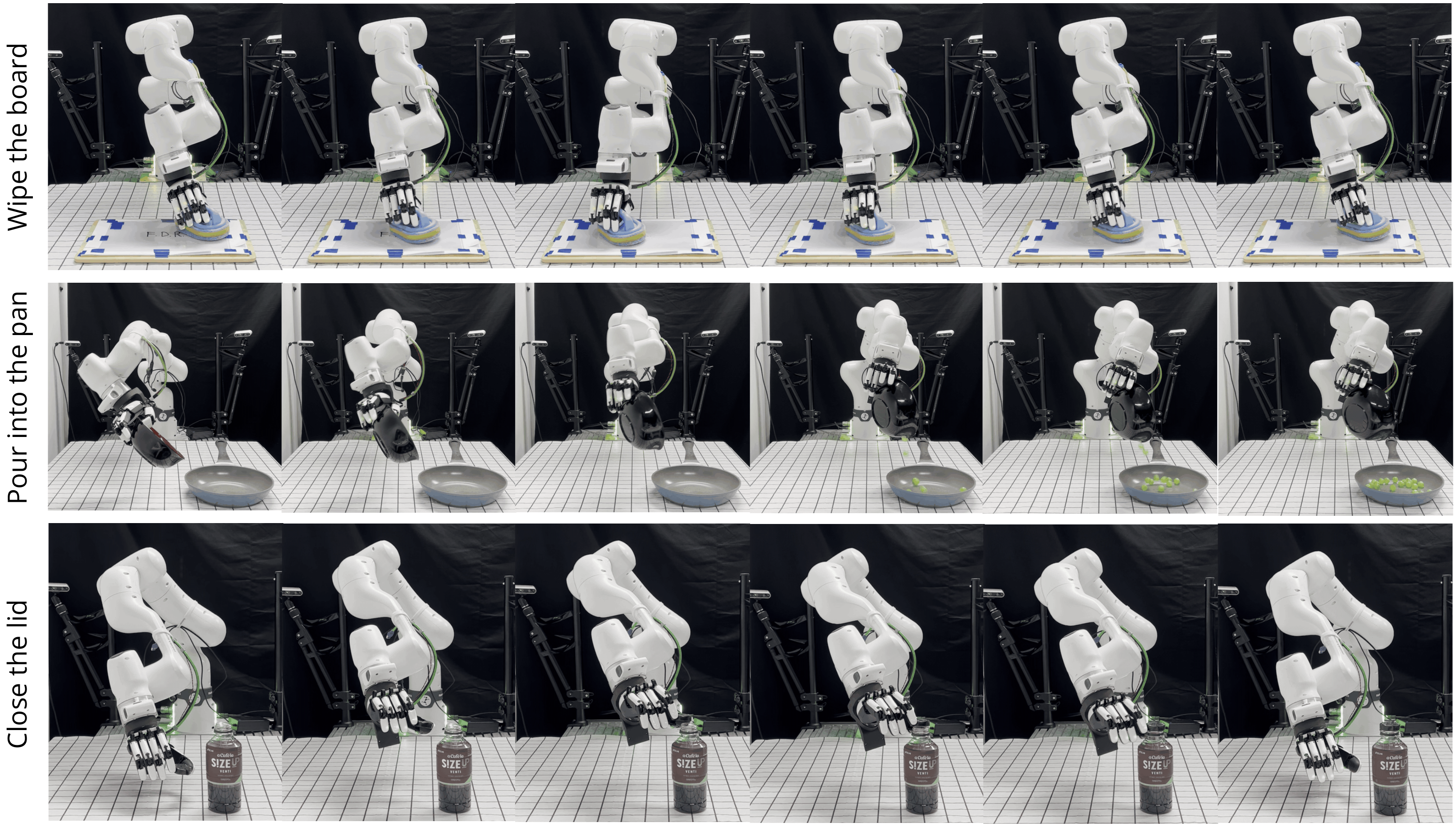}
  \caption{\textbf{Real-world validation of retargeted trajectories.}
The robot successfully executes three retargeted manipulation tasks in the real world: wiping a board, pouring into a pan, and closing a lid.}
  \label{fig:rq5_real_world}
  \vspace{-3mm}
\end{figure*}

\paragraph{Geometry-aware tracking and interaction rewards}
To evaluate the contribution of geometry-aware tracking and \editedgrammar{hand-object}
interaction supervision, we ablate the two components separately and jointly.
In \textit{w/o PCD matching}, we remove the point-cloud observations and
replace the point-cloud object tracking reward with the object tracking reward
used in DexMachina.
In \textit{w/o interaction reward}, we remove the \editedgrammar{hand-object} distance features
$\mathbf{d}^{\mathrm{cur}}_t$ together with the \editedgrammar{hand-object} interaction reward.
We also evaluate a variant that removes both components.
Removing either component reduces learning speed and retargeting
success, while removing both causes the largest degradation.
The larger drop without PCD matching indicates that geometry-aware object
representation and tracking play a larger role in multi-reference retargeting.
At the same time, the interaction reward provides a complementary gain, with
the full model learning fastest and achieving the highest cumulative success.

\paragraph{Hand-wise \texorpdfstring{\editedgrammar{actor-critic}}{actor-critic} architecture}
We compare our hand-wise design with a unified \editedgrammar{actor-critic} that controls
both hands and is trained with their combined reward.
In contrast, our design assigns each hand a separate \editedgrammar{actor-critic} pair
trained with its own reward, while both actors observe the full bimanual state.
With this separation, our method accumulates successful trajectories faster
and achieves higher cumulative retargeting success.
These results support training separate \editedgrammar{actor-critic} pairs with hand-specific
rewards to help each hand perform its role in bimanual manipulation.

\paragraph{Off-policy learning and capacity scaling}
We evaluate the effects of off-policy learning and capacity scaling using
PPO, default FlashSAC, FlashSAC with a larger replay buffer, and our full
configuration with both a larger replay buffer and critic.
Default FlashSAC uses a replay-buffer capacity of 10 million transitions and
a critic hidden dimension of 256. The larger-buffer variant increases the
capacity to 50 million, while our full configuration further increases the
critic hidden dimension to 1024.
This ablation shows the most pronounced performance gap among our design
choices.
Under the same budget of 300 million environment steps, PPO achieves
near-zero cumulative success, whereas default FlashSAC converts approximately
$34\%$ of the reference collection.
Increasing the replay-buffer capacity raises this to approximately $60\%$,
and further scaling the critic reaches $86\%$ with our full configuration.
The large and consistent gains indicate that both effective experience reuse
and sufficient critic capacity are particularly important for learning over
the broad state distribution induced by many reference trajectories.

\begin{figure}[h]
      \centering
      \includegraphics[width=0.8\columnwidth]{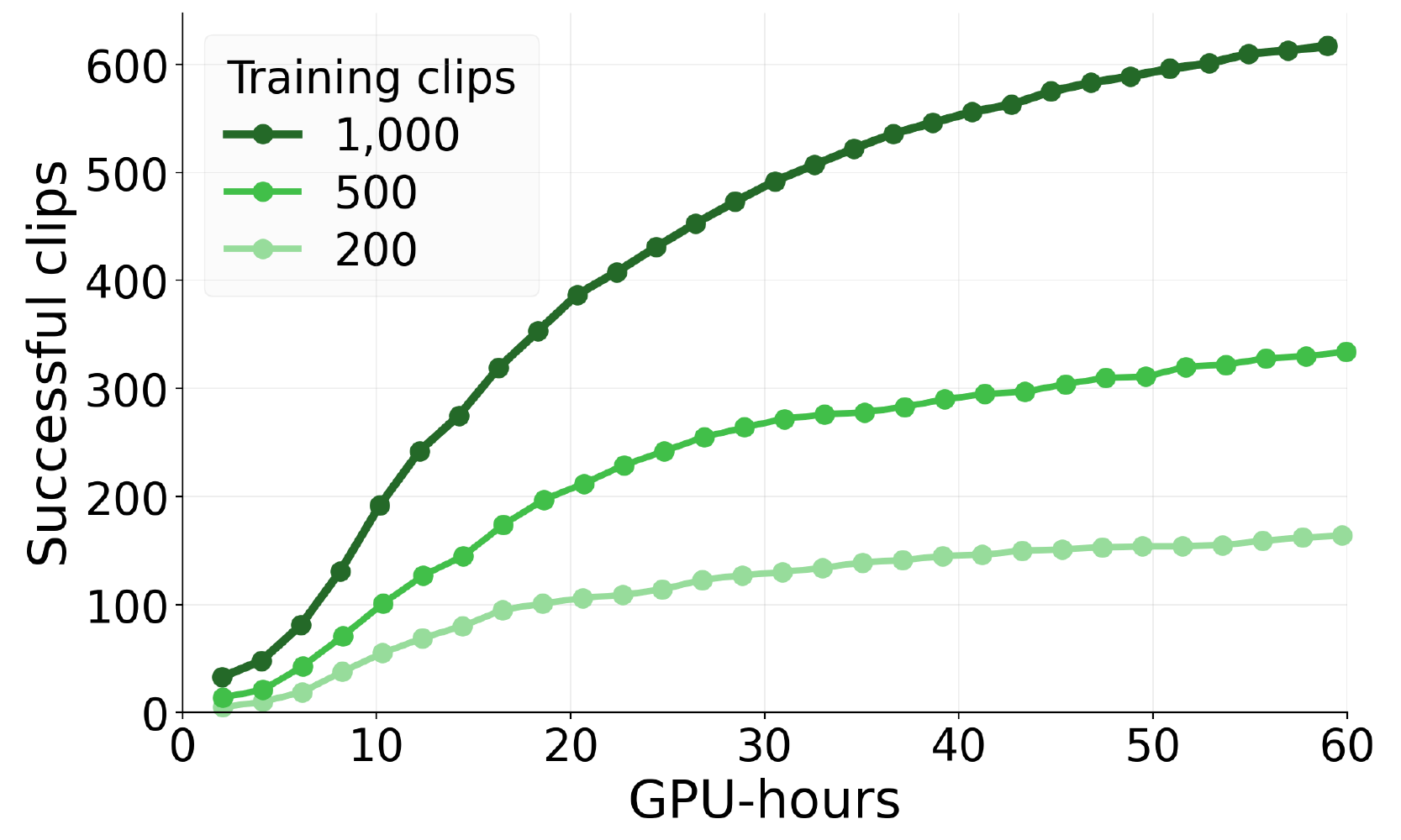}
      \vspace{-2mm}
      \caption{\textbf{Scaling to larger reference collections.}
Cumulative successful conversions for shared policies on 200, 500, and 1,000 reference motions. Larger collections yield more conversions for the same GPU-hour budget.}
      \label{fig:rq3_scaling}
      \vspace{-4mm}
\end{figure}

\subsection{Scaling to Larger Reference Collections}
\label{sec:scaling}

To examine scalability beyond the 50-motion setting, we train shared policies
on collections of 200, 500, and 1,000 reference motions.
We use a budget of 600 million environment steps for all three settings,
providing sufficient interaction as the number of reference motions increases.
To compare their computational efficiency on a common scale,
Figure~\ref{fig:rq3_scaling} shows the cumulative number of successful
conversions over the first 60 GPU-hours of training.
Larger reference collections consistently produce more successful conversions
within the same compute budget, with the gap increasing over the course of
training.
This indicates that exposing a shared policy to more reference motions allows
the computation spent on training to benefit a larger set of demonstrations.
Overall, these results demonstrate that multi-reference training improves
conversion throughput as the reference collection grows.

\subsection{Real-World Validation}

To test whether simulation-retargeted trajectories remain executable under
real-world contact dynamics, we replay successful trajectories on the physical
robot.
As shown in Figure~\ref{fig:rq5_real_world}, the robot completes several
manipulation tasks, including wiping a board, pouring into a pan, and closing
a lid.
These tasks involve different interaction requirements, from sustained surface
contact during wiping to controlled object tilting during pouring and
contact-guided closure.
Successful execution across these tasks demonstrates that the retargeted
trajectories preserve task-relevant \editedgrammar{hand-object} interactions and remain
physically executable beyond simulation.

\section{Conclusion}
\label{sec:conclusion}

We introduced FlashDexRetarget, an RL framework that efficiently converts human hand-object demonstrations into dexterous robot data through multi-reference tracking.
The framework combines geometry- and interaction-aware observations, future reference conditioning, hand-object interaction rewards, and dedicated hand-wise actor-critic networks with off-policy learning and increased replay-buffer and critic capacity.
These components enable experience reuse across demonstrations, achieving higher retargeting success at substantially lower computational cost than the evaluated baselines on XHand and Sharpa Wave Hand.
Ablations demonstrate the contributions of individual components, while scaling experiments show higher conversion throughput with larger reference collections under the same compute budget.
Together, these results support joint retargeting as an efficient approach to generating reusable robot data at scale.

\editedgrammar{Nevertheless,} our framework relies on relatively clean human hand-object trajectories, accurate object geometry, and privileged simulator information, which may limit robustness and real-world deployment. Future work will explore robust reference processing from human videos, generalization to unseen motions and embodiments, task-aware evaluation beyond pose-based metrics, and teacher-student policy distillation to enable deployment with only real-world observations.

\bibliographystyle{IEEEtran}
\bibliography{reference}

\end{document}